# An Empirical Methodology for Detecting and Prioritizing Needs during Crisis Events

M. Janina Sarol, Ly Dinh, Rezvaneh Rezapour, Chieh-Li Chin, Pingjing Yang, Jana Diesner
University of Illinois at Urbana-Champaign, IL, USA
`{mjsarol,dinh4,rezapou2,cchin6,py2,jdiesner}@illinois.edu`

## Abstract

In times of crisis, identifying the essential needs is a crucial step to providing appropriate resources and services to affected entities. Social media platforms such as Twitter contain vast amount of information about the general public's needs. However, the sparsity of the information as well as the amount of noisy content present a challenge to practitioners to effectively identify shared information on these platforms. In this study, we propose two novel methods for two distinct but related needs detection tasks: the identification of 1) a list of resources needed ranked by priority, and 2) sentences that specify *who-needs-what* resources. We evaluated our methods on a set of tweets about the COVID-19 crisis. For task 1 (detecting top needs), we compared our results against two given lists of resources and achieved 64% precision. For task 2 (detecting *who-needs-what*), we compared our results on a set of 1,000 annotated tweets and achieved a 68% F1-score.

## 1 Introduction

In times of crisis, substantial amounts of information about the crisis are shared and discussed on social media platforms (Palen and Anderson, 2016; Vieweg et al., 2010). Some of these posts may contain relevant information about the needs of the affected and at-risk populations (Basu et al., 2018; Dutt et al., 2019; Purohit et al., 2014). The recent COVID-19 virus outbreak is no exception; online platforms such as Twitter are crucial means for sharing information about the global impact of the outbreak (Singh et al., 2020), personal accounts from infected patients (Jimenez-Sotomayor et al., 2020), and situational updates from medical professionals (Rosenberg et al., 2020). Crisis responders and practitioners have also turned to online platforms to obtain actionable information that could aid them in response planning (Vieweg et al., 2010; Zade et al., 2018). In particular, scholars in crisis informatics have provided solutions to detect relevant Twitter messages that express resource needs and availabilities of resources in crisis events such as the 2015 Nepal earthquake (Basu et al., 2017; Dutt et al., 2019) and the 2015 Chennai floods (Sarkar et al., 2019). We build upon and extend this prior literature by proposing two needs detection tasks and applying needs detection to data about the COVID-19 crisis. In particular, we perform (1) *top needs detection* by using word embeddings to extract closest terms to *needs* and *supplies*, and (2) *specific needs detection* to identify entities who need particular resources.

This study makes two contributions. First, we propose a method to identify and prioritize resource needs during a crisis. In addition, we present a set of heuristics to determine the entities that need specific resources. In general, our study serves to provide a reliable set of methods for response professionals to quickly identify immediate areas of needs in the general population and to make effective decisions accordingly.

## 2 Related Work

A large body of literature in crisis informatics has used methods from natural language processing and machine learning to extract relevant situational awareness content from large text corpora (Vieweg et al., 2010; Verma et al., 2011). One of several categories of situational awareness content is needs expressed by affected individuals and communities (Imran et al., 2016; Purohit et al., 2014; **?**; Temnikova et al., 2015). Imran, Mitra, and Castillo (2016) in their analysis of tweets about eight major natural disaster events found that about 21.7% of all tweets contained crucial information about urgent needs for shelter, donations, and essential supplies, such as medical aid, clothing, food, and water. Varga and colleagues (**?**) leveraged machine learning models to match tweets indicating problems and aid being offered to minimize the waste of resources during crisis. Similarly, Purohit and colleagues (2014) classified tweets based on requests and offers of resources, and further matched requests with offers using 18 sets of regular expressions. Temnikova, Castillo, and Vieweg (2015) developed a lexical resource that contained 23 categories of situational awareness, most of which are based on the needs requested and available resources (e.g. clean water, shelter material) as well as services (e.g. rescue workers, relief work) to meet the needs. Basu and colleagues (2017; 2019) identified need and availability tweets, and matched the identified needs with availabilities (Basu et al., 2018). Our paper builds upon this prior work, which primarily focuses on classifying need/non-need tweets, and propose methods that identify a general overview of the top need and specify where and by whom these resources are needed.

## 3 Data

We collected 665,667 tweets from February 28, 2020 to May 8, 2020, with a maximum of 10,000 samples (limit set by firehose) for each day from Crimson Hexagon firehose[1]. Each tweet contains at least one of the hashtags: #COVID19, #COVID-19, #coronavirusoutbreak, #WuhanCoronavirus, #2019nCoV, #CCPvirus, #coronavirus, #CoronavirusPandemic, #SARS-CoV-2, #coronavírus, #wuhanflu, #kungflu, #chineseviruscorona, #ChinaVirus19, #chinesevirus. Our sample includes only tweets from users in the United States and tweets written in English.

## 4 Methodology

### 4.1 Top Needs Detection

For our top needs detection, we trained an embedding model on the dataset, and found the terms closest to the seed terms *needs* and *supplies*. Specifically, we performed the following steps:

1. Detect phrases using AutoPhrase (Shang et al., 2018), setting the threshold for salient phrases to 0.8, and annotate dataset with phrases.

2. Split tweets into sentences and tokens using the NLTK (Loper and Bird, 2002) sentence tokenizer and NLTK TweetTokenizer, respectively.

[1] https://forsight.crimsonhexagon.com/

3. Run word2vec (Mikolov et al., 2013) on the tokenized sentences.

4. Select the top 100 nouns closest to the word embeddings of *needs* and *supplies*.

To identify nouns, we ran the NLTK part-of-speech (POS) tagger on the tweets (prior to phrase annotation). We considered nouns as words whose most frequent POS tag is a noun. A phrase is considered a noun if its final token is a noun (e.g., *testing-capacity* is a noun as *capacity* is a noun).

### 4.2 Specific Needs Detection

We also developed a rule-based method to identify *who needs what* sentences, where *who* is an entity (noun or pronoun) and *what* is a resource or an item (noun). We leveraged the grammatical structure of sentences by using a dependency parser to identify sentences that contain this triple. For this task, we developed two simple rules to identify these types of sentences.

The first rule considers the use of need as a verb. This is a straight-forward application of the *who-needs-what* format. We identified sentences where *who* is the subject and *what* is the direct object. Specifically, after identifying that *need* (or its other word forms) is used as a verb, we selected sentences where the left child of *need* in the dependency parse tree is a nominal subject (nsubj), and the right child is a direct object (dobj). Figure 1 shows an example sentence following the format of this rule along with its dependency parse tree. The second rule considers the use of need as a noun. Our initial data exploration identified many sentences of the format *X is in need of Y*, where in the dependency parse tree, the *who* and *what* are not direct children of the term *need*. The *who* is a child of a copular verb (e.g. is), which is an ancestor of *need*. The term linking the copular verb and *need* is a preposition (i.e. the copular verb is the term's parent and *need* is its prepositional object (pobj). The *what* is a descendant of need, also linked through a preposition. Figure 2 shows an example sentence

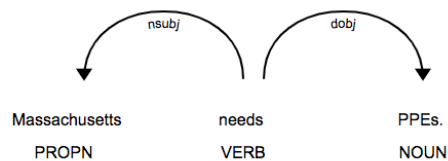

Figure 1: Rule considering need as a verb

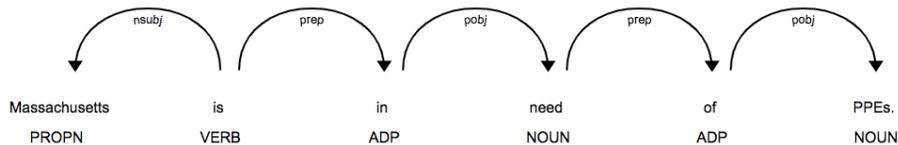

Figure 2: Rule considering need as a noun

following the format of this rule and its dependency parse tree.

Similar to the top needs detection task, we also used the NLTK sentence tokenizer and NLTK TweetTokenizer to split the tweets into sentences and tokens, respectively. We used spaCy (Honnibal and Montani, 2017) to generate the dependency parse trees. Our source codes for both top and specific needs detection are available on GitHub (repository link redacted for review).

### 4.3 Evaluation

There is no single comprehensive list of resources needed by people in the U.S. for the COVID-19 crisis. However, we found two sets of sources that we deemed as proxies for such a list. First, the World Health Organization's (WHO) essential resource planning guidelines (2020) provide a set of forecasting tools and documents for calculating the required manpower, supplies, and equipment needed to adequately respond to the virus. Second, the U.S. Department of Health and Human Services (HHS) Office of Inspector General published the results of a survey conducted about hospitals' experiences in responding to the pandemic (Grimm, 2020). To evaluate our results for the top needs detection task, we counted the number of matches between the list that we had generated with the resources that appeared in the WHO guidelines and HHS survey. This helps to capture precision. We report our results as precision@k, with k ranging from 10 to 100.

For the specific needs detection task, two annotators identified *who-needs-what* sentences from a random set of 1,000 sentences that contain any word form of need (i.e., need, needs, needing, and needed). Each annotator was assigned 600 sentences, where 200 sentences also appeared in another annotator's list. Cohen's kappa is 0.91.

We report our results on the specific needs detection task using precision, recall, and F1-score. We compare our work to the needs detection algorithm proposed by Basu and colleagues (Basu et al., 2017), which classified need and non-need tweets by ranking tweets based on their cosine similarities to the embeddings of the terms *need* and *requir* (stemmed). We set the cut-off value of the need-related tweets to 250 and performed the same preprocessing steps outlined in the paper. While their work is focused on identifying all need tweets, this is the closest work to our particular task.

## 5 Results

Table 1 shows the top 10 resources generated by our top needs detection method. The full set of results is shown in the appendix. Comparing with the WHO guidelines, the precision@10 is 0.8, and comparing with the HHS survey, it is 0.9. When both WHO and HHS documents are considered, the precision@10 is 1. The top 13 terms (and 19 of the top 20 terms) appear in at least either one of the WHO or HHS documents. Overall, 41 of the 100 terms appear in the WHO guidelines, 57 in the HHS survey, and 64 in at least one document.

Figure 3 shows the precision@k, where k is in increments of 10. There is a steep drop-off in the results when the cut-off is relaxed from 20 to 30, but after this drop-off, the precision@k decreases in a more controlled rate. This indicates that resources needed still appear in the lower parts of the list. High precision scores for lower k values suggest that our proposed top needs detection method is not only able to identify resources needed, but also to produce a rigorous ranking of needs.

| Resource | WHO | HHS |
|---|---|---|
| medical-equipment | ✓ | ✓ |
| equipment | ✓ | ✓ |
| medical-supplies | ✓ | ✓ |
| protective-gear | ✓ | ✓ |
| stockpile | ✗ | ✓ |
| protective-equipment | ✓ | ✓ |
| ppe | ✓ | ✓ |
| manufacturing | ✗ | ✓ |
| personal-protective-equipment | ✓ | ✓ |
| medicines | ✓ | ✗ |

Table 1: Top Resources Generated for COVID-19

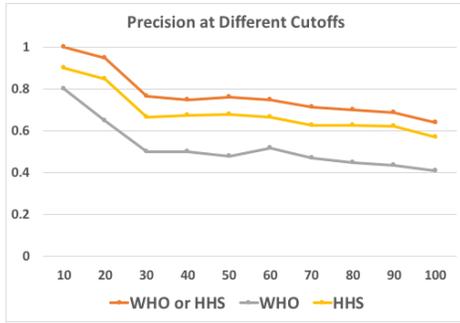

Figure 3: Precision at Different Cutoffs

For the specific needs detection task, our method produced a precision of 0.66, recall of 0.70, and F1-score of 0.68. Sentences that were incorrectly predicted as positive examples (i.e., *who-needs-what*) include those of the form *if you need x, then..*, while false negatives include more complex sentences. Only using the first rule produces precision of 0.66, recall of 0.68, and an F1-score of 0.67; indicating that most *who-needs-what* sentences follow this rule, where the *who* is the subject of the sentence or clause and the *what* is the direct object. Our baseline method based on the work by Basu and colleagues (2017) performed poorly, achieving only 0.28 precision, 0.26 recall, and 0.27 F1-score.

## 6 Discussion

The top needs detection results vary in terms of specificity (e.g., equipment vs. medical equipment, personal protective equipment vs. respirators, medical supplies vs. essential items). Several of the terms that are not on the WHO and HHS lists are general terms such as goods, aid, efforts, programs, and assets. In addition, several terms are synonymous (e.g., personal protective equipment and PPE). This suggests that clustering the terms may lead to a more distinct set of results.

It is not surprising that more of the terms we detected appeared in the HHS than in the WHO document because we collected our tweet data from the U.S., and the HHS document is from a survey of U.S. hospitals, while the WHO list has a more global audience. Overall, our results suggest two findings: 1) our top needs detection method works, and 2) most COVID-19 needs mentioned on Twitter are either of medical or financial nature.

Our specific needs detection results show that a simple rule-based method can identify sentences that mention entities needing resources and the resources needed. Further, a single rule is able to identify 68% of the sentences. This is an interesting finding with several implications. For one, we can produce a simple white-box method for identifying *who-needs-what* sentences. While the use of deep learning may increase results, our method requires no training data. Another implication is that it suggests that the mention of needs on social media, particularly Twitter, is mostly uniform and follows a specific format; this uniformity could be due to the limited characters available in a tweet. Testing the generalizability of this method to other crisis events is part of our future work.

While social media has been shown to be a valuable source of information during crises, finding useful information is still akin to finding a needle in a haystack. For our specific needs detection task, we only found 262 positive examples (out of 1,000 sample sentences containing a word form of need).

Our method is able to generate a ranked set of needs for 600,000+ tweets in less than 30 minutes. Running steps such as phrase detection and POS tagging in parallel may even improve this time. For the specific needs detection task, our method can classify 1,000 sentences in 8 seconds.

## 7 Conclusion and Future Work

In our work, we presented two needs detection methods: one to identify a list of top resources needed during a crisis, and another one to specifically identify the entities who need resources. We believe that the combination of these two methods are necessary to capture the wide range of needs present during crisis events. Specific to the COVID-19 crisis, our results suggest that there are still a number of unmet needs of the general population and affected stakeholders, most of which are about the lack of protective equipment and financial assistance. It is imperative for practitioners and associated stakeholders to be aware of these needs to properly plan and respond effectively.

In future work, we aim to expand our method to also specifically identify availabilities, including whether the needs have been met, and entities who met the needs. In addition, we plan to identify a more comprehensive set of requests, including *hopes*, *wants*, and *wishes* during a crisis.

| | | | |
|---|---|---|---|
| medical-equipment | materials | hand-sanitizer | grants |
| equipment | access | face-masks | relief |
| medical-supplies | demand | gloves | essential-workers |
| protective-gear | essential-goods | local-hospitals | capability |
| stockpile | production | respirators | groceries |
| protective-equipment | face-shields | healthcare-workers | devices |
| ppe | personnel | recipients | pharmacies |
| manufacturing | federal-funding | refused | flexibility |
| personal-protective-equipment | reagents | essential-supplies | masks |
| medicines | federal-assistance | barriers | living-wage |
| #ppe | ventilators | demands | national-stockpile |
| supply | systems | repairs | medical-facilities |
| distribution | assets | relief-funds | assistance |
| goods | capacity | food-banks | packages |
| manufacturers | programs | utilities | trace |
| funds | aid | meds | dpa |
| plans | economic-relief | testing-capacity | purchases |
| essentials | kits | defense-production-act | handouts |
| essential-items | gowns | childcare | machines |
| financial-relief | food | ability | deliveries |
| needing | funding | services | local-governments |
| necessities | efforts | providers | paid-sick-leave |
| critical-supplies | medication | requirements | shortages |
| clean-water | supply-chain | surgical-masks | failed |
| resources | facilities | expenses | hospitals |

Table A1: Top 100 Resources Generated for COVID-19